\pdfoutput=1

\documentclass[11pt]{article}

\usepackage[]{ACL}
\usepackage{graphicx}
\usepackage{times}
\usepackage{latexsym}
\usepackage{todonotes}
\usepackage{multirow}
\usepackage{amssymb}
\usepackage[inline]{enumitem}
\usepackage{amsthm}
\usepackage{dsfont}
\usepackage{relsize}
\usepackage{subcaption}
\usepackage[T1]{fontenc}

\usepackage[utf8]{inputenc}

\usepackage{microtype}

\usepackage{inconsolata}
\usepackage{subcaption}

\title{Quantifying Misattribution Unfairness in Authorship Attribution}

\author{
Pegah Alipoormolabashi \\
Stony Brook University \\
\texttt{palipoormola@cs.stonybrook.edu}
\And
Ajay Patel \\
University of Pennsylvania \\
\texttt{ajayp@seas.upenn.edu}
\AND
Niranjan Balasubramanian \\
Stony Brook University \\
\texttt{niranjan@cs.stonybrook.edu}
}

\begin{document}

\newcommand{\pegah}[2][]{\todo[inline, color=blue!30, #1]{#2}}
\newcommand{\nb}[2][]{\todo[inline, color=green!30, #1]{#2}}
\newcommand{\ajay}[2][]{\todo[inline, color=red!30, #1]{#2}}

\newcommand{\newstuff}[2]{\textcolor{red}{#2}}

\newcommand{\luar}[0]{LUAR}
\newcommand{\mpnet}[0]{MPNet\textsubscript{AR}}
\newcommand{\mpnetst}[0]{MPNet\textsubscript{st}}
\newcommand{\sbert}[0]{SBERT}
\newcommand{\wegmann}[0]{Wegmann}
\newcommand{\styledistance}[0]{StyleDist.}

\newcommand{\ufar}[1]{MAUI\textsubscript{#1}}
\newcommand{\indicator}[1]{\mathds{1}\left[#1\right]}

\newcommand{\rot}[1]{\begin{turn}{90}#1\enspace\end{turn}}
\newcommand{\mr}[1]{\multicolumn{1}{|c|}{#1}}
\maketitle
\begin{abstract}
Authorship misattribution can have profound consequences in real life. In forensic settings simply being considered as one of the potential authors of an evidential piece of text or communication can result in undesirable scrutiny. This raises a fairness question: Is every author in the candidate pool 
at equal risk of misattribution? Standard evaluation measures for authorship attribution systems do not explicitly account for this notion of fairness. We introduce a simple measure, Misattribution Unfairness Index (\ufar{k}), which is based on how often authors are ranked in the top $k$ for texts they did \emph{not} write. Using this measure we quantify the unfairness of five models on two different datasets. All models exhibit high levels of unfairness with increased risks for some authors. Furthermore, we find that this unfairness relates to how the models embed the authors as vectors in the latent search space. In particular, we observe that the risk of misattribution is higher for authors closer to the centroid (or center) of the embedded authors in the haystack. These results indicate the potential for harm and the need for communicating with and calibrating end users on misattribution risk when building and providing such models for downstream use.

\end{abstract}

\section{Introduction}

Authorship attribution has various sensitive uses in multiple domains such as literature \citep{zhao2007searching,stanczyk2007machine} and forensics ~\citep{introtoforensic2016}. In some uses, authorship misattribution or even suspected authorship can have dire consequences for the individuals involved. For example, consider a scenario where a forensic investigation hinges on finding who wrote a piece of text from a large pool of potential authors. Automatic authorship attribution is often seen as a way to reduce the number of candidates for such settings~\citep{tschuggnall-etal-2019-reduce}. 
However, even being suspected of authorship in this setting can result in further scrutiny. 
Given such undesirable consequences, we ask a question of fairness: \emph{Are some authors more likely to be misattributed than others?}

To answer this question, we study the commonly used \emph{needle-in-the-haystack} formulation~\citep{rivera-soto-etal-2021-learning,Wang2023CanAR,Wen2024AIDBenchAB}. In this setting, the \emph{haystack} is a collection of known authors and documents they have authored $D_h$. Given some  documents $D_q$ from a query author i.e., ones whose authorship is unknown as of yet, the attribution task is to find a candidate author (\emph{needle}) from the haystack who most likely authored the query documents. The task is then framed as a ranking problem, where systems rank haystack authors by comparing an embedding of their documents $e_i = \mbox{enc}(D_{h}(a_i))$ to an embedding of the query documents $e_q = \mbox{enc}(D_q)$. Such \emph{embed-and-rank} solutions are appealing since they allow efficient scaling to large haystacks with many authors~\cite{douze2024faiss}. 

Existing evaluation measures for these solutions do not tackle fairness notions. Prior work mostly use effectiveness metrics, such as Mean Reciprocal Rank (MRR), or Recall at various ranks (R@k), or Recall at various ranks (R@k) ~\citep{andrews2019learninginvariantrepresentationssocial,khan-etal-2021-deep,rivera-soto-etal-2021-learning,10.1007/978-3-642-00887-0_61}. 
These metrics are designed to reduce the amount of manual scrutiny by assessing whether the correct author is ranked as highly as possible. They do not capture or measure the risks of getting ranked highly for other authors. 

We make two contributions to remedy this gap: \textbf{1)} We introduce a way to measure the unfairness in rankings induced by models and empirically assess unfairness across five models over two datasets. \\ \textbf{2)} We perform an analysis and provide a potential explanation for unfairness in how embedding distributions relate to  misattribution risks of authors. Our results show that model rankings exhibit high levels of unfairness and authors closer to the center of the embedding space are at a higher risk for misattribution. These call for further research both in evaluation and in modeling of authorship attribution systems to reduce the potential for unfairness-related harms.

\section{Misattribution Unfairness in Author Ranking}
\label{unfairness-define}
We introduce a notion of fairness where all users carry equal risks of being misattributed i.e., are equally likely to be ranked high for documents that they have not authored. This is similar to the common definition of fairness in retrieval settings \citep{biega2020overviewtrec2019fair}.  In retrieval, the focus is that all authors receive relevance-proportional attention in the rankings. Here we focus on reducing undue or disproportional presence in ranking.

Suppose there are $N_{h}$ authors in the haystack and $N_{q}$ query authors selected at random from the haystack. For any single query, if we consider an unbiased ranking i.e. a random permutation of the authors, the probability of any specific author being ranked higher than $k$ is $\frac{k}{N_{h}}$. When we query for $N_{q}$ times, an author is expected to get ranked higher than $k$ for $E_k =\lceil\frac{k}{N_{h}} \times N_{q}\rceil$ times. 

The unfairness of model-induced rankings can thus be characterized in terms of how the actual counts of authors being ranked at top $k$ exceeds the expected count\footnote{We acknowledge that expecting a random permutation does not factor in the demographic or stylistic ``relevance'' of non-query authors to the query author. For example, an author who shares a regional dialect with the query author is more relevant to the query in this regard. While this notion of relevance is useful for attribution, it can be unfair in forensic settings as if a member of Demographic X is a suspect of a crime, undue scrutiny should not be brought to all members of Demographic X. While imperfect, a random permutation is a reasonable calibration to benchmark systems against and measure the magnitude of bias towards certain authors.}. Let $c^k_j$ denote the number of times author $a_j$ is ranked in the top $k$. Then, we can quantify the unfairness as follows:
\setlength{\abovedisplayskip}{3pt}
\begin{equation}
\label{ufar}
\centering
\mbox{\ufar{k}} = \frac{\sum_{j=1}^{N_h} \max(0 , (c^k_j - E_k ))}{ k \times (N_q - E_k)}
\end{equation}
This metric normalizes the sum of the differences by its highest possible value which happens in the worst case: when the same $k$ authors are ranked in top $k$ for all queries. This scales the values between 0 and 1, 0 being most fair and 1 being the least. 



\section{Evaluation}
\label{sec:experiments}

We experiment with multiple authorship models and datasets and use cosine similarity for ranking.

\subsection{Experimental Setup}
\paragraph{Datasets} 
(i) Reddit: We use the evaluation partition of the dataset by \citet{andrews2019learninginvariantrepresentationssocial} with 111,396 candidate authors and 25,000 query authors. (ii) Bloggers: We select 9000 bloggers from the authorship corpus of Blogger posts \citep{Schler2006EffectsOA} and use 2500 of them as queries. (iii) Fanfiction: We randomly select 20,000 fanfiction authors, and use 7500 of them as queries.

\paragraph{Models} We use five text embedding models:
\begin{enumerate*}
    \item \sbert{} \citep{reimers2019sentencebertsentenceembeddingsusing}: A sentence transformer model based on DistilRoBERTa \citep{Sanh2019DistilBERTAD}
    \item \luar{} \citep{rivera-soto-etal-2021-learning}: A universal authorship embedding model trained on the Reddit Million User Dataset \citep{khan-etal-2021-deep}.
    \item Style Embedding \cite{wegmann-etal-2022-author}: A sentence transformer built on RoBERTa-base \citep{liu2019robertarobustlyoptimizedbert} and trained for style representation. We call this model \wegmann.
    \item \styledistance{} ~\citep{patel2024styledistancestrongercontentindependentstyle}: Another style embedding model trained on a combination of Reddit comments and synthetic data.
    \item \mpnet{}: Microsoft's sentence transformer (\mpnetst) by \citet{song2020mpnetmaskedpermutedpretraining} that we train for authorship representation. 
\end{enumerate*} Section \ref{apx:models-datasets} of the appendix provides more details on models and datasets.

We show the performance of the five embedding models on the two datasets in Table~\ref{tab:basic-scores}. In the rest of this section we look at the models' embeddings and rankings from other angles.
\begin{table*}[h!]
\centering
\begin{tabular}{c|cc|cc|cc|}
\cline{2-7}
 & \multicolumn{2}{c|}{Blogs} & \multicolumn{2}{c|}{Reddit} & \multicolumn{2}{c|}{Fanfiction} \\ \hline
\multicolumn{1}{|c|}{Model} & \multicolumn{1}{c|}{R@8} & MRR & \multicolumn{1}{c|}{R@8} & MRR  & \multicolumn{1}{c|}{R@8} & MRR \\ \hline
\multicolumn{1}{|c|}{\sbert} & \multicolumn{1}{c|}{0.61} & 0.48 & \multicolumn{1}{c|}{0.15} & 0.10 & \multicolumn{1}{c|}{0.28} & 0.22 \\ \hline
\multicolumn{1}{|c|}{\luar} & \multicolumn{1}{c|}{0.97} & 0.90 & \multicolumn{1}{c|}{0.82} & 0.71 & \multicolumn{1}{c|}{0.53} & 0.44\\ \hline
\multicolumn{1}{|c|}{\mpnet} & \multicolumn{1}{c|}{0.96} & 0.88 & \multicolumn{1}{c|}{0.40} & 0.30 & \multicolumn{1}{c|}{0.30} & 0.25 \\ \hline
\multicolumn{1}{|c|}{\wegmann} & \multicolumn{1}{c|}{0.45} & 0.32 & \multicolumn{1}{c|}{0.08} & 0.05 & \multicolumn{1}{c|}{0.09} & 0.06\\ \hline
\multicolumn{1}{|c|}{\styledistance} & \multicolumn{1}{c|}{0.68} & 0.55 & \multicolumn{1}{c|}{0.09} & 0.06  & \multicolumn{1}{c|}{0.16} & 0.12 \\ \hline
\end{tabular}
\caption{Recall-at-8 and Mean Reciprocal Rank scores of embedding models on three datasets.}
\label{tab:basic-scores}
\end{table*}

\begin{table}[ht]
\begin{subtable}{\linewidth}
\centering
\begin{tabular}{c|cccc|}
\cline{2-5}
 & \multicolumn{4}{c|}{k} \\ \hline
\multicolumn{1}{|c|}{Model} & \multicolumn{1}{c|}{5} & \multicolumn{1}{c|}{10} & \multicolumn{1}{c|}{15} & 20 \\ \hline
\multicolumn{1}{|c|}{\sbert} & \multicolumn{1}{c|}{0.20} & \multicolumn{1}{c|}{0.31} & \multicolumn{1}{c|}{0.36} & 0.39 \\ \hline
\multicolumn{1}{|c|}{\luar} & \multicolumn{1}{c|}{0.06} & \multicolumn{1}{c|}{0.12} & \multicolumn{1}{c|}{0.16} & 0.18 \\ \hline
\multicolumn{1}{|c|}{\mpnet} & \multicolumn{1}{c|}{0.09} & \multicolumn{1}{c|}{0.17} & \multicolumn{1}{c|}{0.22} & 0.25 \\ \hline
\multicolumn{1}{|c|}{\wegmann} & \multicolumn{1}{c|}{0.03} & \multicolumn{1}{c|}{0.09} & \multicolumn{1}{c|}{0.13} & 0.15 \\ \hline
\multicolumn{1}{|c|}{\styledistance} & \multicolumn{1}{c|}{0.07} & \multicolumn{1}{c|}{0.15} & \multicolumn{1}{c|}{0.19} & 0.22 \\ \hline
\end{tabular}
\caption{Reddit}
\label{tab:unfair-reddit-scores}
\vspace{0.1cm}
\end{subtable}
\begin{subtable}{\linewidth}
\centering
\begin{tabular}{c|cccc|}
\cline{2-5}
 & \multicolumn{4}{c|}{k} \\ \hline
\multicolumn{1}{|c|}{Model} & \multicolumn{1}{c|}{5} & \multicolumn{1}{c|}{10} & \multicolumn{1}{c|}{15} & 20 \\ \hline
\multicolumn{1}{|c|}{\sbert} & \multicolumn{1}{c|}{0.24} & \multicolumn{1}{c|}{0.36} & \multicolumn{1}{c|}{0.37} & 0.40 \\ \hline
\multicolumn{1}{|c|}{\luar} & \multicolumn{1}{c|}{0.15} & \multicolumn{1}{c|}{0.26} & \multicolumn{1}{c|}{0.27} & 0.31 \\ \hline
\multicolumn{1}{|c|}{\mpnet} & \multicolumn{1}{c|}{0.12} & \multicolumn{1}{c|}{0.23} & \multicolumn{1}{c|}{0.23} & 0.27 \\ \hline
\multicolumn{1}{|c|}{\wegmann} & \multicolumn{1}{c|}{0.06} & \multicolumn{1}{c|}{0.14} & \multicolumn{1}{c|}{0.13} & 0.17 \\ \hline
\multicolumn{1}{|c|}{\styledistance} & \multicolumn{1}{c|}{0.11} & \multicolumn{1}{c|}{0.22} & \multicolumn{1}{c|}{0.21} & 0.25 \\ \hline
\end{tabular}
\caption{Blogs}
\label{tab:unfair-blog-scores}
\end{subtable}
\vspace{0.1cm}
\begin{subtable}{\linewidth}
\centering
\begin{tabular}{c|cccc|}
\cline{2-5}
 & \multicolumn{4}{c|}{k} \\ \hline
\multicolumn{1}{|c|}{Model} & \multicolumn{1}{c|}{5} & \multicolumn{1}{c|}{10} & \multicolumn{1}{c|}{15} & 20 \\ \hline
\multicolumn{1}{|c|}{\sbert} & \multicolumn{1}{c|}{0.21} & \multicolumn{1}{c|}{0.30} & \multicolumn{1}{c|}{0.34} & 0.36 \\ \hline
\multicolumn{1}{|c|}{\luar} & \multicolumn{1}{c|}{0.17} & \multicolumn{1}{c|}{0.25} & \multicolumn{1}{c|}{0.28} & 0.30 \\ \hline
\multicolumn{1}{|c|}{\mpnet} & \multicolumn{1}{c|}{0.17} & \multicolumn{1}{c|}{0.25} & \multicolumn{1}{c|}{0.28} & 0.30 \\ \hline
\multicolumn{1}{|c|}{\wegmann} & \multicolumn{1}{c|}{0.08} & \multicolumn{1}{c|}{0.12} & \multicolumn{1}{c|}{0.14} & 0.15 \\ \hline
\multicolumn{1}{|c|}{\styledistance} & \multicolumn{1}{c|}{0.12} & \multicolumn{1}{c|}{0.18} & \multicolumn{1}{c|}{0.21} & 0.22 \\ \hline
\end{tabular}
\caption{Fanfiction}
\label{tab:unfair-fanfict-scores}
\end{subtable}

\caption{(\ufar{k}) for different $k$ values across models and datasets. \ufar{k} is defined in section \ref{unfairness-define} as a measure of an authorship attribution system's unfairness in misattributing authors. The scores range between 0 (most fair) and 1 (most unfair).} 

\label{tab:unfair-scores}
\end{table}
\renewcommand{\arraystretch}{1}
\subsection{Misattribution Unfairness in Model Rankings}
We measure \ufar{k} (eq. \ref{ufar}) for different values of $k$ and show the results in Table~\ref{tab:unfair-scores}. Higher values show higher unfairness. Putting Table~\ref{tab:unfair-scores} and \ref{tab:basic-scores} together, we can see that there is no clear relationship between how good a model is in correctly attributing authors and how fair it is in misattributing authors.  
For example, \wegmann{} is the worst performing model in terms of R@8 and MRR, but it consistently shows the least unfairness. \mpnet{} and \styledistance{} have very close \ufar{} scores, yet their ranking performance varies greatly. Even \luar{} with very high ranking scores does not guarantee unfairness in misattribution. In fact, after \sbert, \luar{} is the most unfair model to the Bloggers.

\renewcommand{\arraystretch}{1.1}

\begin{table}[ht]
\begin{subtable}{\linewidth}
\centering
\smaller
\begin{tabular}{c|c|c|c|}
\cline{2-4}
 & $> 2 \times E_{10}$ & $> 4 \times E_{10}$ & $> 5 \times E_{10}$ \\ \hline
 \multicolumn{1}{|c|}{\sbert} & 8487 & 2582 & 1599 \\ \hline
\multicolumn{1}{|c|}{\luar} & 4290 & 242 & 54 \\ \hline
\multicolumn{1}{|c|}{\mpnet} & 6054 & 701 & 299 \\ \hline
\multicolumn{1}{|c|}{\wegmann} & 2967 & 18 & 3 \\ \hline
\multicolumn{1}{|c|}{\styledistance} & 5411 & 382 & 106 \\ \hline
\end{tabular}
\caption{Reddit}
\label{tab:num-exceed-reddit}
\vspace{0.1cm}
\end{subtable}
\begin{subtable}{\linewidth}
\centering
\smaller
\begin{tabular}{c|c|c|c|}
\cline{2-4}
 & $> 2 \times E_{10}$ & $> 4 \times E_{10}$ & $> 5 \times E_{10}$ \\ \hline
 \multicolumn{1}{|c|}{\sbert} & 858 & 296 & 214 \\ \hline
\multicolumn{1}{|c|}{\luar} & 821 & 189 & 96 \\ \hline
\multicolumn{1}{|c|}{\mpnet} & 816 & 122 & 56 \\ \hline
\multicolumn{1}{|c|}{\wegmann} & 496 & 9 & 1 \\ \hline
\multicolumn{1}{|c|}{\styledistance} & 789 & 104 & 33 \\ \hline
\end{tabular}
\caption{Blogs}
\label{tab:num-exceed-blogs}
\end{subtable}

\begin{subtable}{\linewidth}
\centering
\smaller
\begin{tabular}{c|c|c|c|}
\cline{2-4}
 & $> 2 \times E_{10}$ & $> 4 \times E_{10}$ & $> 5 \times E_{10}$ \\ \hline
 \multicolumn{1}{|c|}{\sbert} & 1319 & 328 & 104 \\ \hline
\multicolumn{1}{|c|}{\luar} & 1161 & 170 & 38 \\ \hline
\multicolumn{1}{|c|}{\mpnet} & 1155 & 181 & 43 \\ \hline
\multicolumn{1}{|c|}{\wegmann} & 316 & 0 & 0 \\ \hline
\multicolumn{1}{|c|}{\styledistance} & 780 & 32 & 0 \\ \hline
\end{tabular}
\caption{Fanfiction}
\label{tab:num-exceed-fanfic}
\end{subtable}
\caption{Number of authors who are ranked in top 10 more than the expected number i.e. $E_{10}$. Note that the size of the haystack and the number of the queries are different across datasets, hence the huge difference in the numbers between the three tables.}
\label{tab:num-exceed}
\end{table}
\renewcommand{\arraystretch}{1}


\begin{table}[!ht]
\centering
\begin{tabular}{lccc}
\hline
\textbf{Model} & \textbf{Reddit} & \textbf{Blogs} & \textbf{Fanfic} \\
\hline
LUAR      & 9.75  & 10.0 & 12.0 \\ 
SBERT     & 39.00 & 21.75 & 5.8 \\
Wegmann   & 4.50  & 4.25&  3.6\\
MPNet     & 12.25 & 10.25 & 12.0 \\
StyleDist & 8.25  & 7.0 &  19.6\\
\hline
\end{tabular}
\caption{Extreme cases of unfair misattribution risk. Numbers show the ratio of the number of times an author is ranked in top 10 to $E_{10}$ for the author bearing the highest risk of misattribution. }
\label{extremes}
\end{table}
To present the unfairness from another viewpoint, we count the number of authors who carry higher risks of being ranked in top $k = 10$. Tables \ref{tab:num-exceed-reddit} and \ref{tab:num-exceed-blogs} show how these counts compare to the expected count $E_{10}$. The number of unfairly misattributed authors and the severity of this issue vary across models. Similar to the trends in Table~\ref{tab:unfair-scores} \wegmann{} is the most fair of the models on both Reddit and Blogs datasets. Again, for Blogs, \luar{} is highly unfair despite its superior performance. Looking at the last column of Table~\ref{tab:num-exceed-blogs}, \luar's number of misattributed authors is more than \mpnet, \styledistance, and \wegmann{} together.
Table~\ref{tab:num-exceed} shows the counts of the number of authors subject to unfair misattribution. The differences between columns show that the extent of this misattribution risk varies for different authors. In section \ref{apx:risk_across_authors} of the appendix we show how the risk is distributed among authors. Looking at the authors with the most misattribution we see that for instance, with \sbert{} there is -at least- one author for whom the misattribution risk is almost 40x the random case.



\subsection{Relation to Embedding Distribution}
How the author embeddings are distributed is central to the effectiveness of attribution and fairness of misattribution in rankings. One way to analyze this relationship is via the distance of author embeddings to the centroid (center) of the embeddings. 

\begin{figure}[ht!]
\centering
        \begin{subfigure}[b]{0.475\textwidth}  
            \centering 
            \includegraphics[width=\textwidth]{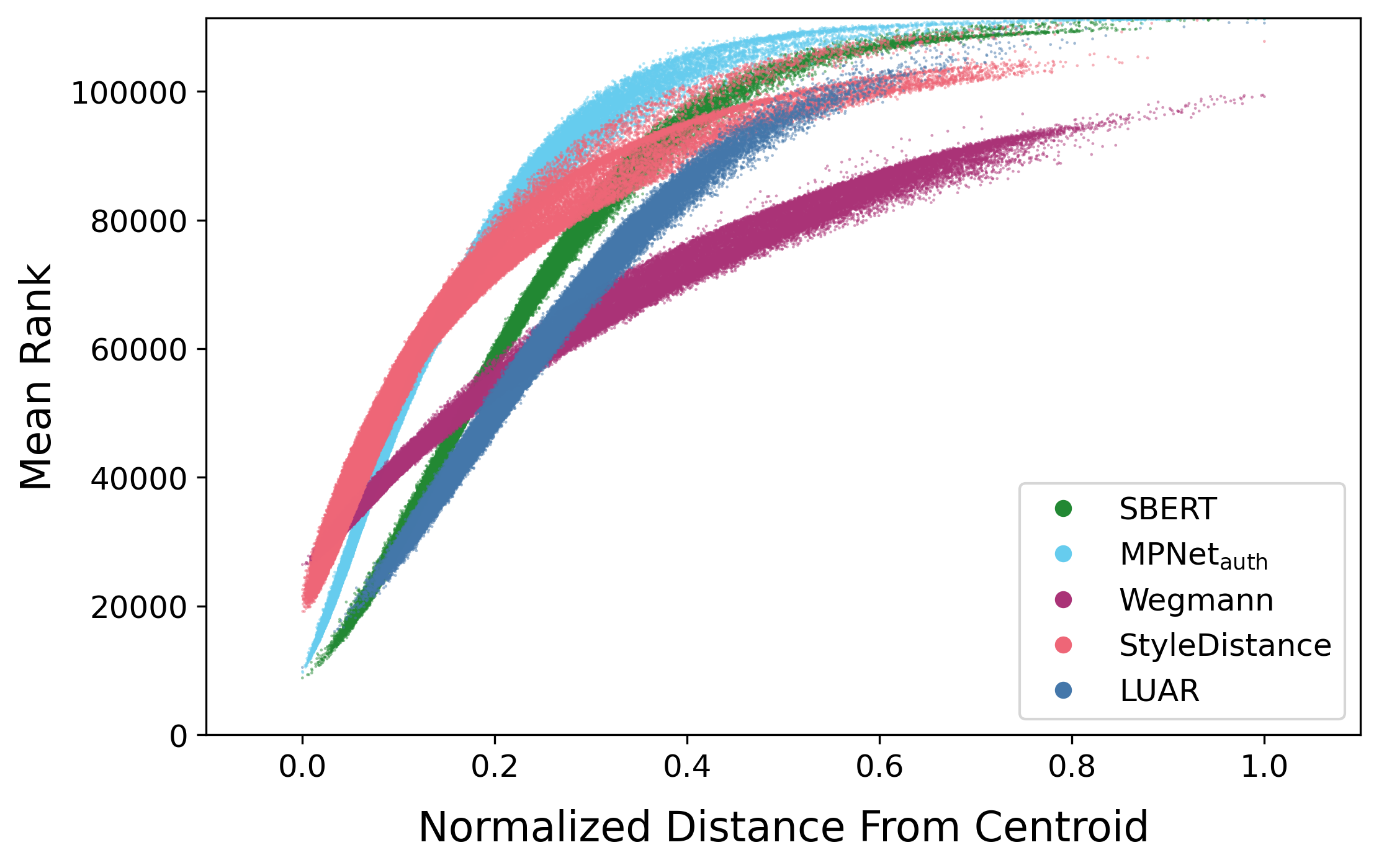}
            \caption{Reddit}   
        \end{subfigure}
        \vspace{0.3cm}
        \begin{subfigure}[b]{0.475\textwidth}   \centering
            \centering 
            \includegraphics[width=\textwidth]{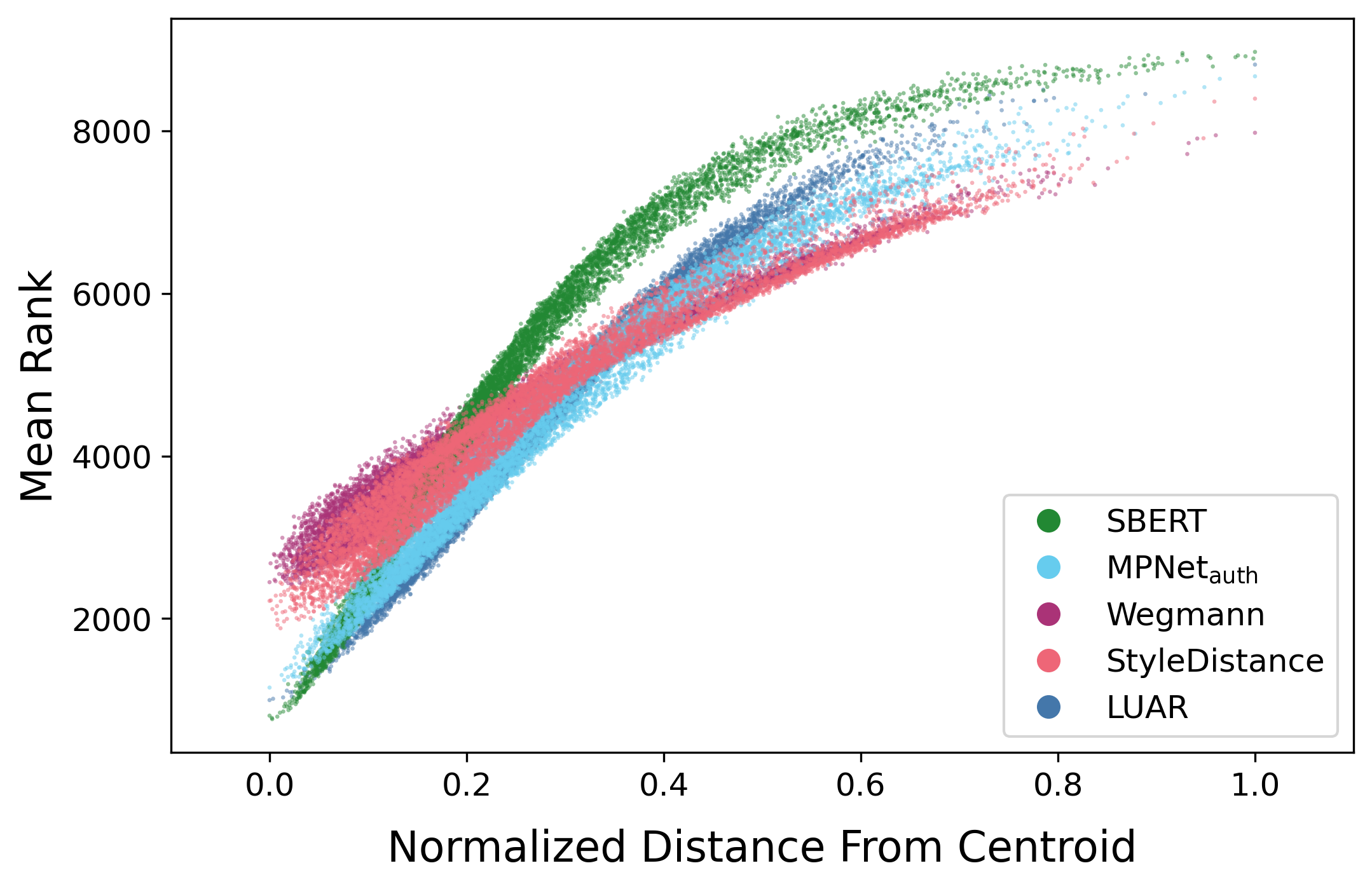}
            \caption{Blogs} 
        \end{subfigure}
                \vspace{0.3cm}
        \begin{subfigure}[b]{0.475\textwidth}   \centering
            \centering 
            \includegraphics[width=\textwidth]{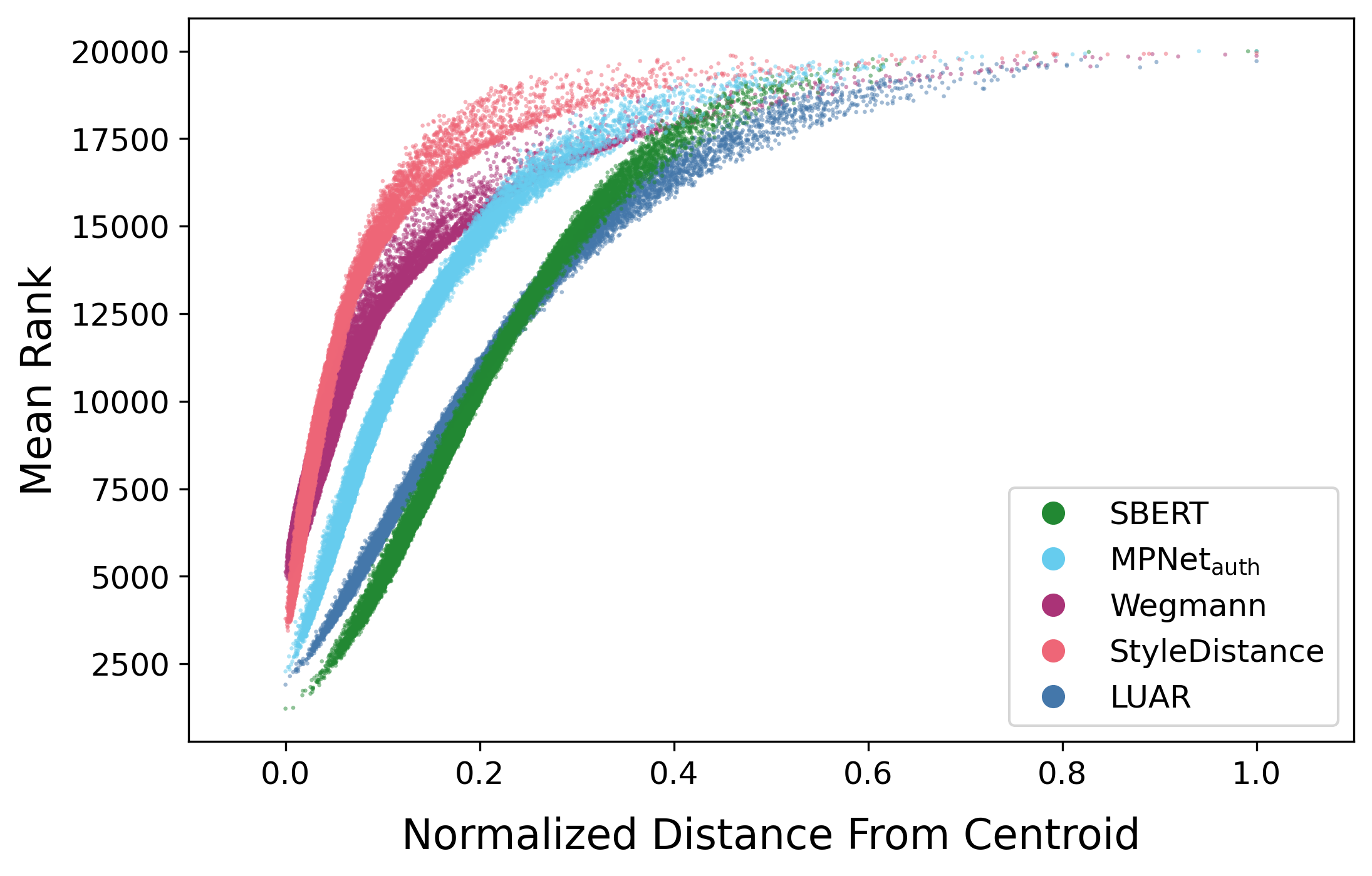}
            \caption{Fanfiction} 
        \end{subfigure}
        \caption{Relationship between authors' average rank and their distance from the centroid. Distances are min-max normalized. Authors' mean rank is highly correlated with their distance from centroid.}
        \label{fig:meanrank-distance}
\end{figure}

To this end, we average all author embeddings to compute the centroid and measure distance of each author to the centroid as $1 - cosine$. We plot authors' mean rank (average of their ranking across all queries) against their distance to centroid in figure \ref{fig:meanrank-distance}. For better visualization the distance values are scaled to the [0,1] range. 
We see strong correlations between authors' distance to the centroid and their risk of misattribution, as measured by their average rank over all queries. 
Across models and datasets the closer an author is to the centroid, the lower their average rank is; i.e. authors closer to the centroid are ranked higher on average. Each author is ranked for queries that are chosen at random, therefore  an author's average ranking should not correlate with their distance from the centroid. We expect the plots to be close to a horizontal line. The expected average rank for every author is the middle of the ranked list i.e., half the size of haystack. Among the models we compare, \wegmann{} curves are closest to the ideal. Note that these correlations depend largely on how the embeddings are distributed. Indeed, we find that distance distributions vary considerably across different models. See Figure \ref{fig:distance-distribution} in Appendix Section \ref{apx:distance-distribution}.

\subsection{Unfairness of missed attribution}
\label{section-missed-attribution}
The centroid analyses can also help us understand which authors are harder to find i.e., ones who are not recognized as the author of their own documents.  To assess how often an author is ranked high for their own documents, we average their reciprocal ranks over multiple query subsets drawn from their documents\footnote{Specifically, we perform four queries per each query author, each comprising four of their documents.}. Authors that are ranked close to the top when queried have higher MRRs than those who are not.

We then test three hypotheses relating MRR to authors' distance from centroid: (i) Authors with higher MRR are further away from the centroid compared to authors with lower MRR, (ii) Authors with higher MRR are further away from the centroid compared to a random subset of authors. (iii) Authors with lower MRR are closer to the center than a random subset of authors. We use the Mann-Whitney U test \footnote{This test does not assume normality.} \citep{Mann1947OnAT} to accept or reject our hypotheses. Table~\ref{tab:test-stats} in Appendix shows statistics for all tests. The numbers show that for Reddit the first and the second hypotheses are statistically supported for all models and the third only for \luar. For blogs, only the third hypothesis for \sbert{} is rejected. See figures \ref{fig:needle-dist} and \ref{fig:blog-needle-dist} for visualizations of this phenomenon. To summarize, while authors closer to centroid are more likely to be ranked higher when queried for other authors' documents, they are not necessarily ranked high for their own. These results also demonstrate the importance of embedding distributions to the fairness of authorship rankings. 




\section{Related Work}
Automatic authorship analysis has a rich body of work~\citep[see][]{huang2025authorshipattributionerallms,tyo2022stateartauthorshipattribution}, with a heavy emphasis on usefulness and reliability of features used in authorship analysis (e.g. \citet{chaski2001empirical, baayen2002experiment}, or the effects of the attribution setup (e.g. \citet{Stamatatos2013OnTR, Sari2018TopicOS}). While these provide general insights into authorship attribution, to the best of our knowledge, there have been no specific studies that focus on notions of fairness in authorship attribution. 
Fairness of NLP models has been studied in many application domains including dialogue \citep{Liu2019DoesGM}, language modeling \citep{Cao2022OnTI,Qian2022PerturbationAF}, text generation \citep{Fleisig2023FairPrismEF}, classification \cite{Pruksachatkun2021DoesRI} and clinical NLP \citep{Meng2022InterpretabilityAF}. \citet{gallegos2024bias} survey research on bias and fairness in large language models focusing on metrics, probing datasets, and bias mitigation techniques. There have also been extensive research on fairness and bias related metrics. \citet{czarnowska-etal-2021-quantifying} survey, categorize, and compare some of the fairness and bias related metrics.

The closest to our work is research on fairness of embedding-based document retrieval systems.
Their main concerns are: group fairness \citep{yang2016measuringfairnessrankedoutputs}, individual fairness \cite{Biega2018EquityOA, GarcaSoriano2021MaxminFairRI}, their trade-off with each other and with relevance \citep{Gao2019HowFC}, and recently unfairness in ranking with LLMs \citep{wang-etal-2024-large}.
 These works aim to ensure relevant documents from certain individuals or groups are fairly represented in the top ranks. In contrast, in our authorship setting, the focus is on the risks of being included in the top ranks in forensic or law enforcement settings.

\section{Conclusion}
Authorship attribution carries profound risks in settings such as forensic and law enforcement uses. In these settings, measures of effectiveness alone are not adequate for evaluating and understanding the impact of misattribution. To this end, we argued that we also need to understand if the rankings induced by attribution models distribute misattribution risks equally. The unfairness measure we introduced helps quantify these risks. Our empirical measurements of models used to produce authorship embeddings shows that in an embed-and-rank approach there are many authors who are at a substantially higher risk of being ranked in the top $k$. We further showed that this risk correlates with how the embeddings are distributed. Our findings call for careful consideration of such notions of misattribution fairness in evaluation, development, and deployments of authorship analysis systems.

\section*{Limitations}
In practice, the anticipated ``fair'' ranking of the haystack authors is not entirely random, but rather a function of authors' relatedness to the query. The distribution of queries therefore impacts the ``most fair'' baseline, and consequently the unfairness measurements. This impact is not accounted for in this paper. Our results hold for the set of queries selected for in our experiments, which were chosen at random. It is possible different selection methods of queries may impact unfairness measurements.
Additionally, this work's focus is limited to cases where over-attribution is undesired. A broader study would also cover cases where under-attribution is problematic.
\section*{Ethical Considerations}
\paragraph{Potential Misuse} Results and analyses presented in this paper are meant to prompt researchers in authorship attribution to focus on unfairness, assessing it, and preventing its potential harms. People and institutions using authorship attribution may misuse these results to justify automatically preempting groups of candidate authors. We do not investigate characteristics of authors who are easy or difficult to track down. Nevertheless, a malicious agent may reproduce our experiments, analyze hard-to-find authors (near-centroid authors), and use characteristics of their writing as an authorship obfuscation method. 
\paragraph{Our Work} We report licenses and sources of every artifact we used in Appendix \ref{apx:technical}. We did not investigate individuals or use demographic information (e.g. reported age and gender published in the Blogs corpus) about authors in any of our experiments and operate under the assumption that no author should have significantly greater than random expectation of appearing in the top $k$ results when they are not the query author of interest regardless of demographics.

\section*{Acknowledgments}
This research is supported in part by the Office of the Director of National Intelligence (ODNI), Intelligence Advanced Research Projects Activity (IARPA), via the HIATUS Program contract \#2022-22072200005. The views and conclusions contained herein are those of the authors and should not be interpreted as necessarily representing the official policies, either expressed or implied, of ODNI, IARPA, or the U.S. Government. The U.S. Government is authorized to reproduce and distribute reprints for governmental purposes notwithstanding any copyright annotation therein.


\bibliography{ref}
\bibliographystyle{acl_natbib}
\newpage
\appendix

\section{Technical Details of Experiments}
\label{apx:technical}
\subsection{Models and Datasets}
\label{apx:models-datasets}
\renewcommand{\arraystretch}{1.2}

\begin{table*}[t]
\smaller
\begin{tabular}{|c|c|c|c|c|c|}
\hline
\multirow{2}{*}{Model} & \multirow{2}{*}{Base Sentence Transformer} & \multirow{2}{*}{Training Objective} & \multirow{2}{*}{Training Data} & \multirow{2}{*}{Number of Parameters} & \multirow{2}{*}{License} \\
 &  &  &  &  &  \\ \hline
 \sbert & DistilRoBERTa & - & - & 82M & apache-2.0 \\ \hline
\luar & paraphrase-distilroberta-base-v1 & Batch Contrastive Loss* & Reddit (MUD) & 82M & apache-2.0 \\ \hline
\mpnet{}& all-mpnet-base-v2 &\begin{tabular}[c]{@{}c@{}}Multiple-Negative \\ Ranking Loss\end{tabular}  & Reddit(MUD) & 109M & apache-2.0 \\ \hline
\wegmann{}& RoBERTa-base & Triplet Loss & Reddit & 124M & mit \\ \hline
\styledistance & RoBERTa-base & Contrastive Loss & \begin{tabular}[c]{@{}c@{}}Reddit + \\ Synthetic Data\end{tabular} & 124M & mit \\ \hline
\end{tabular}
\caption{Overview of embedding models we use in experiments. We perform inference on these models using an NVIDIA TITAN Xp GPU. }
\label{tab:models-detail}
\end{table*}
\renewcommand{\arraystretch}{1}
\subsubsection*{Blogs Data}
 We use the corpus of Blogger posts collected by \citet{Schler2006EffectsOA} as it is commonly used in authorship analysis research. This dataset "may be freely used for non-commercial research purposes."\footnote{See https://u.cs.biu.ac.il/~koppel/BlogCorpus.htm} We filter for authors with more than 10 and fewer than 200 posts. From those we pick 9000 candidate authors, 2500 of which are used as queries and needles. Each author in the haystack has 16 blog posts (randomly selected) and each author in the queries has 10. 
 \subsubsection*{Reddit Data}
  For evaluation we use the evaluation partition of the Reddit dataset created by \citet{andrews2019learninginvariantrepresentationssocial}. It has 111,396 candidate authors and 25,000 query authors. Each author in the haystack has 16 Reddit comments, and each author in the queries has 16 Reddit comments as well. For finetuning \mpnet{} we use a part of the Million User Dataset by \citet{khan-etal-2021-deep}. 
\subsubsection*{Fanfiction Data}
We use a subset of the fanfiction dataset from PAN19 \citep{Kestemont2019OverviewOT} and PAN20 \citep{Kestemont2020OverviewOT} for evaluation. This subset consists of 20000 haystack authors, 7500 of whom are used as queries.
\subsection{Finetuning \mpnetst}
We choose to finetune \mpnetst{} \citep{song2020mpnetmaskedpermutedpretraining} because it has a different structure than the BERT-based models. Also, all-mpnet-base-v2 \footnote{https://huggingface.co/sentence-transformers/all-mpnet-base-v2} is a top-performing sentence transformer. Out of 12 layers we keep 8 frozen for training. This model is licensed under the apache-2.0 license.
We use cached multiple-negative ranking loss implemented in the sentence-transformers library\footnote{https://www.sbert.net/index.html}. Training with multiple negatives has shown to be very effective in the case of \luar{} \cite{rivera-soto-etal-2021-learning}. We set the learning rate to $5e^{-5}$ with a linear scheduler. Maximum sequence length is set to 512, and we train with a batch size of 200. We train for 5000 steps on the training subset of Reddit data by \citet{khan-etal-2021-deep}. Training computation was done on an NVIDIA RTX A6000 GPU. 
\label{apx:mpnet-train}

\subsection{Indexing and Retrieving Embeddings}
We use Faiss \citep{douze2024faiss}\footnote{https://github.com/facebookresearch/faiss} for efficient indexing of author embeddings and performing fast searches. All the indexing and search computations are done on CPU. Faiss library is under an MIT-license. 

\subsection{Libraries and Packages}
Here is a list of major python libraries we use (with no particular order):
\begin{itemize}[nosep]
    \item torch==2.3.1
    \item sentence-transformers==3.3.1
    \item transformers==4.45.2
    \item tensorflow-datasets==4.9.6
    \item scikit-learn==1.5.0
    \item scipy==1.14.0
    \item numpy==1.26.4
    \item nltk==3.8.1
    \item matplotlib==3.9.2
    \item seaborn==0.13.2
    \item faiss-cpu==1.8.0.post1
\end{itemize}
\section{Embedding Distributions}
\label{apx:distance-distribution}
Since authorship attribution in this setting is based entirely upon embeddings and their distances, we use a simple distance-to-centroid measure to characterize the distribution of author embeddings under different models. Intuitively, if all author embeddings are clustered close to each other then the distances to the centroid will be small. If they are spread out then the distances to the centroid will be large. Per each dataset and model we find the centroid of all haystack author embeddings, then calculate each author's distance from this centroid. The distance measure is $1 - cosine$, so it can take values from 0 to 2. 

\begin{figure}[t!]
    \centering
        \begin{subfigure}[b]{0.475\textwidth}
            \centering
            \includegraphics[width=\textwidth]{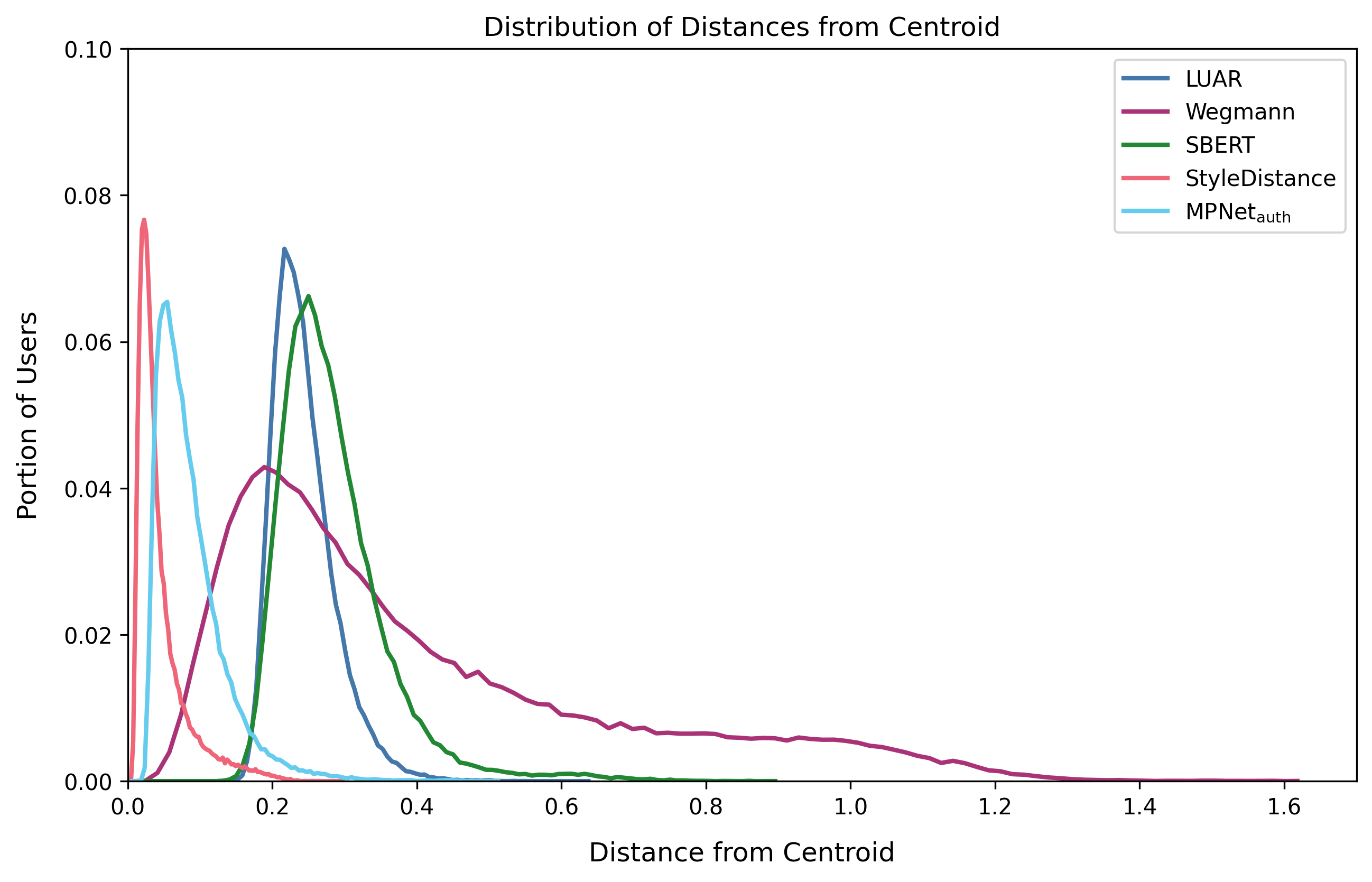}
            \caption{Reddit}  
            \label{fig: dd-reddit}
        \end{subfigure}
        \begin{subfigure}[b]{0.475\textwidth}   
            \centering 
            \includegraphics[width=\textwidth]{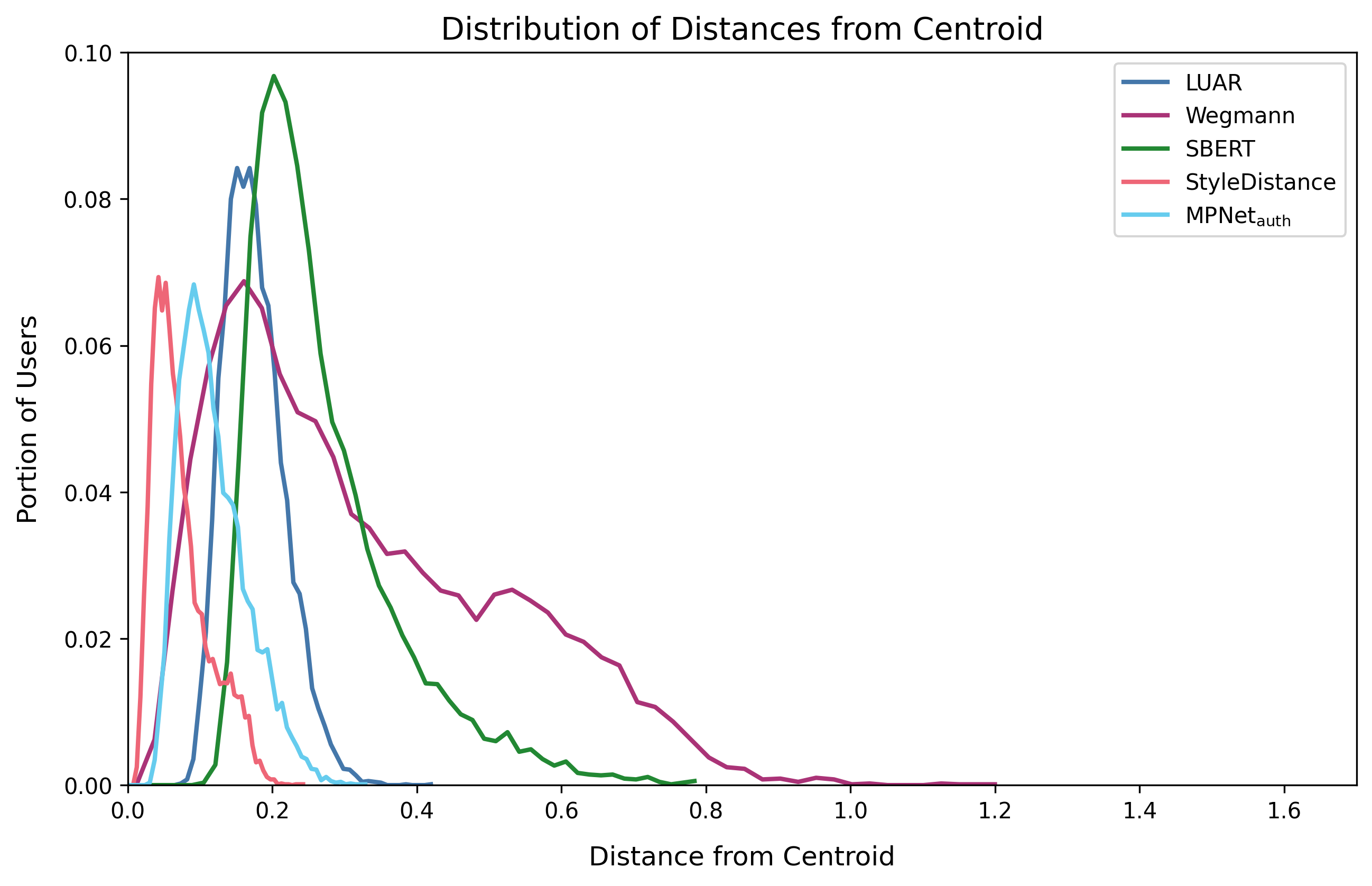}
            \caption{Blogs}   
            \label{fig: dd-blogs}
        \end{subfigure}
\caption{Distribution of authors' distinctness (i.e. their distance from the centroid computed as: $1 - cosine$)}
\label{fig:distance-distribution}
\end{figure}

Figure \ref{fig:distance-distribution} shows the distribution of the distance value per each model for Reddit and Blogs datasets. The diversity of the distributions in terms of domain and curve shape inspired us to investigate how it affects the effectiveness and fairness of authorship attribution. 

\begin{table}[!ht]
\begin{subtable}{\linewidth}
\centering
\begin{tabular}{lccc}
\hline
\textbf{Model} & \textbf{Max} & \textbf{Mean} & \textbf{Std} \\
\hline
LUAR      & 9.75  & 1.62 & 0.50 \\
SBERT     & 39.00 & 2.37 & 1.67 \\
Wegmann   & 4.50  & 1.50 & 0.32 \\
MPNet     & 12.25 & 1.79 & 1.74 \\
StyleDist & 8.25  & 1.69 & 0.56 \\
\hline
\end{tabular}
\caption{Reddit}
\end{subtable}
\vspace{0.2cm}
\begin{subtable}{\linewidth}
\centering

    \begin{tabular}{lccc}
\hline
\textbf{Model} & \textbf{Max} & \textbf{Mean} & \textbf{Std} \\
\hline
LUAR      & 10.0  & 2.07 & 1.10 \\
SBERT     & 21.75 & 2.61 & 2.12 \\
Wegmann   & 4.25  & 1.58 & 0.40 \\
MPNet     & 10.25 & 1.92 & 0.85 \\
StyleDist & 7.00  & 1.84 & 0.70 \\
\hline
\end{tabular}
\caption{Blogs}
\end{subtable}
\vspace{0.2cm}
\begin{subtable}{\linewidth}
\centering

    \begin{tabular}{lccc}
\hline
\textbf{Model} & \textbf{Max} & \textbf{Mean} & \textbf{Std} \\
\hline
LUAR      & 12.0  & 2.03 & 1.05 \\
SBERT     & 19.6& 2.32 & 1.58\\
Wegmann   & 3.6  & 1.54 & 0.39 \\
MPNet     & 12.0 & 2.04 & 1.09 \\
StyleDist & 5.80  & 1.75 & 0.64 \\
\hline
\end{tabular}
\caption{Fanfiction}
\end{subtable}
\caption{Variation of the risk of unfair misattribution across different authors measured by the ratio of the number of times an author is ranked in top 10 to $E_{10}$.}
\label{tab:risk-across-authors}
\end{table}
\section{Analysis of Needle Authors' Distance to Centroid}
\label{apx:needle-analysis}
\begin{figure}
    \centering
        \begin{subfigure}[b]{\linewidth}
            \centering
            \includegraphics[width=\linewidth]{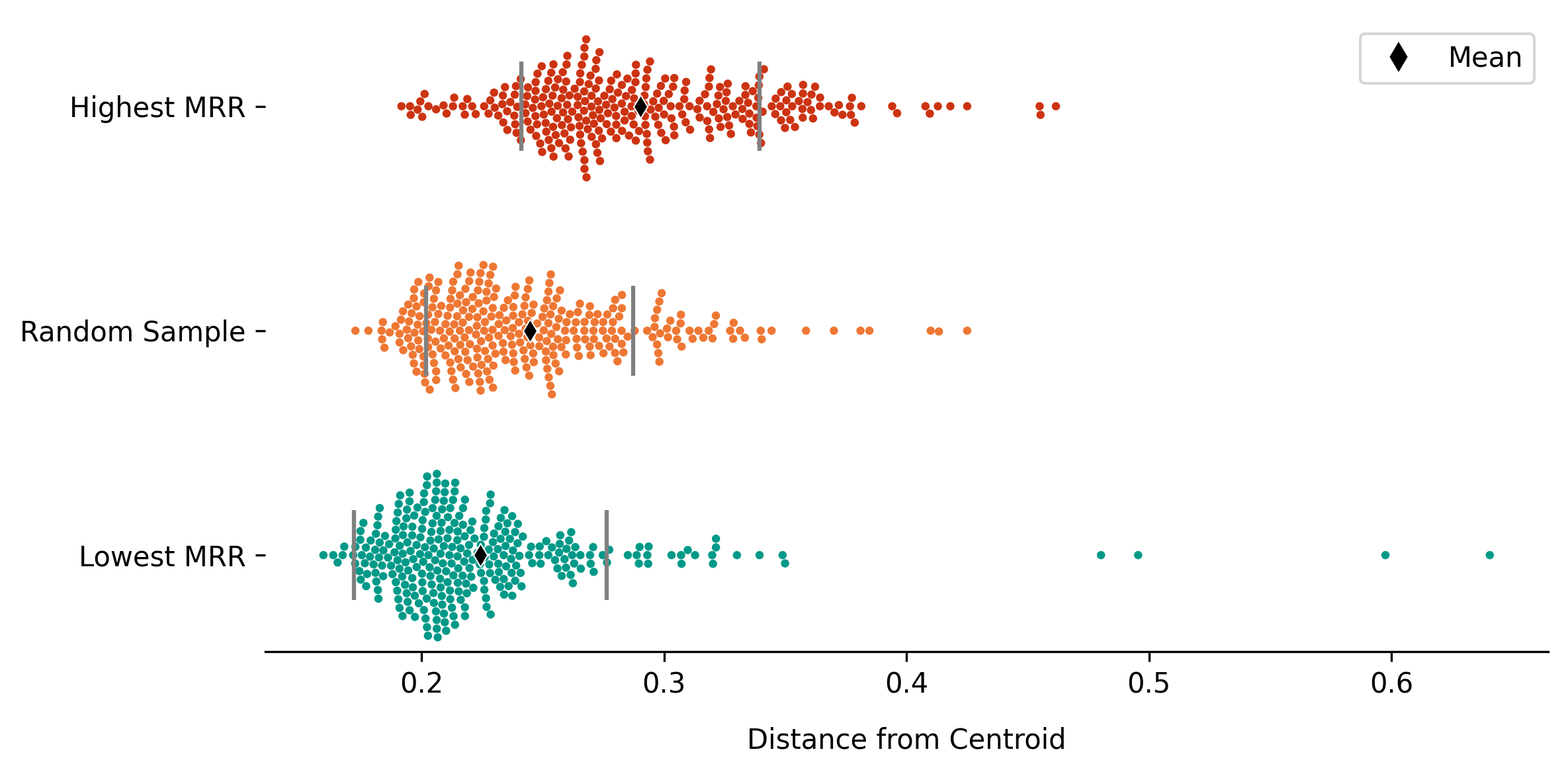}
            \caption{\luar}
        \end{subfigure}
        \begin{subfigure}[b]{\linewidth}
            \centering 
            \includegraphics[width=\linewidth]{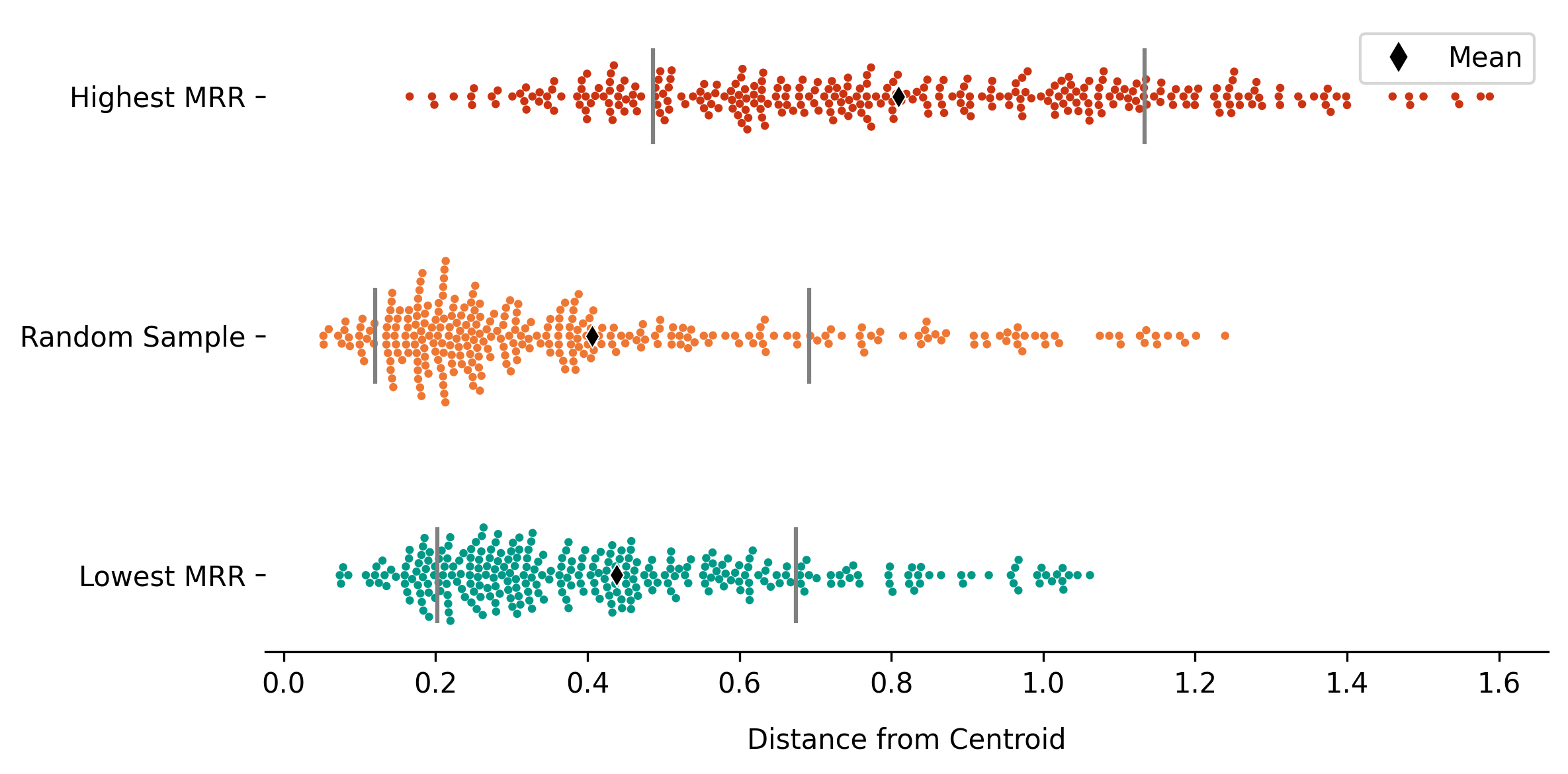}
            \caption{\wegmann}
        \end{subfigure}
        \begin{subfigure}[b]{\linewidth}   
            \centering 
            \includegraphics[width=\linewidth]{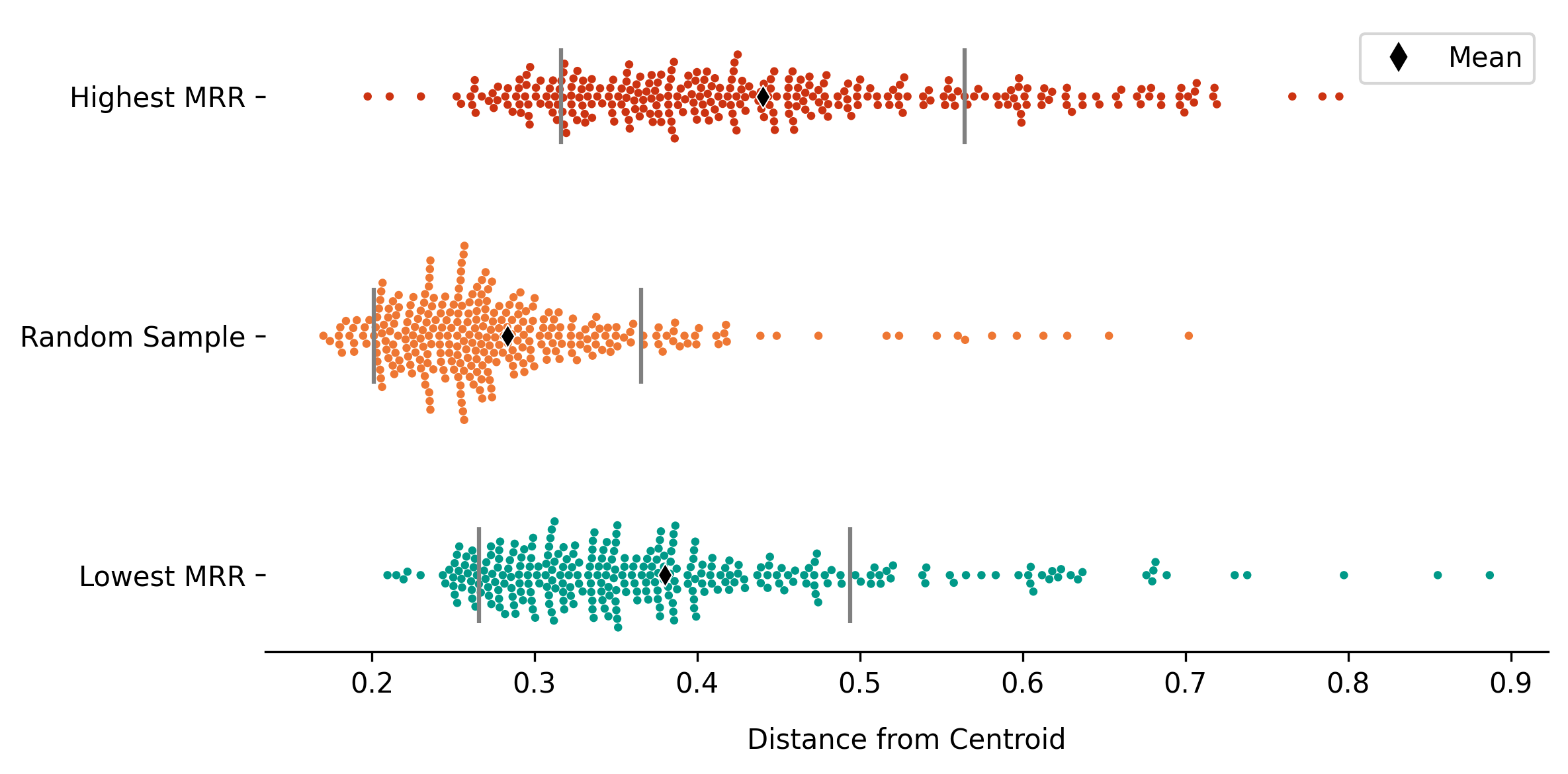}
            \caption{\sbert}
        \end{subfigure}
        \begin{subfigure}[b]{\linewidth}   
            \centering 
            \includegraphics[width=\linewidth]{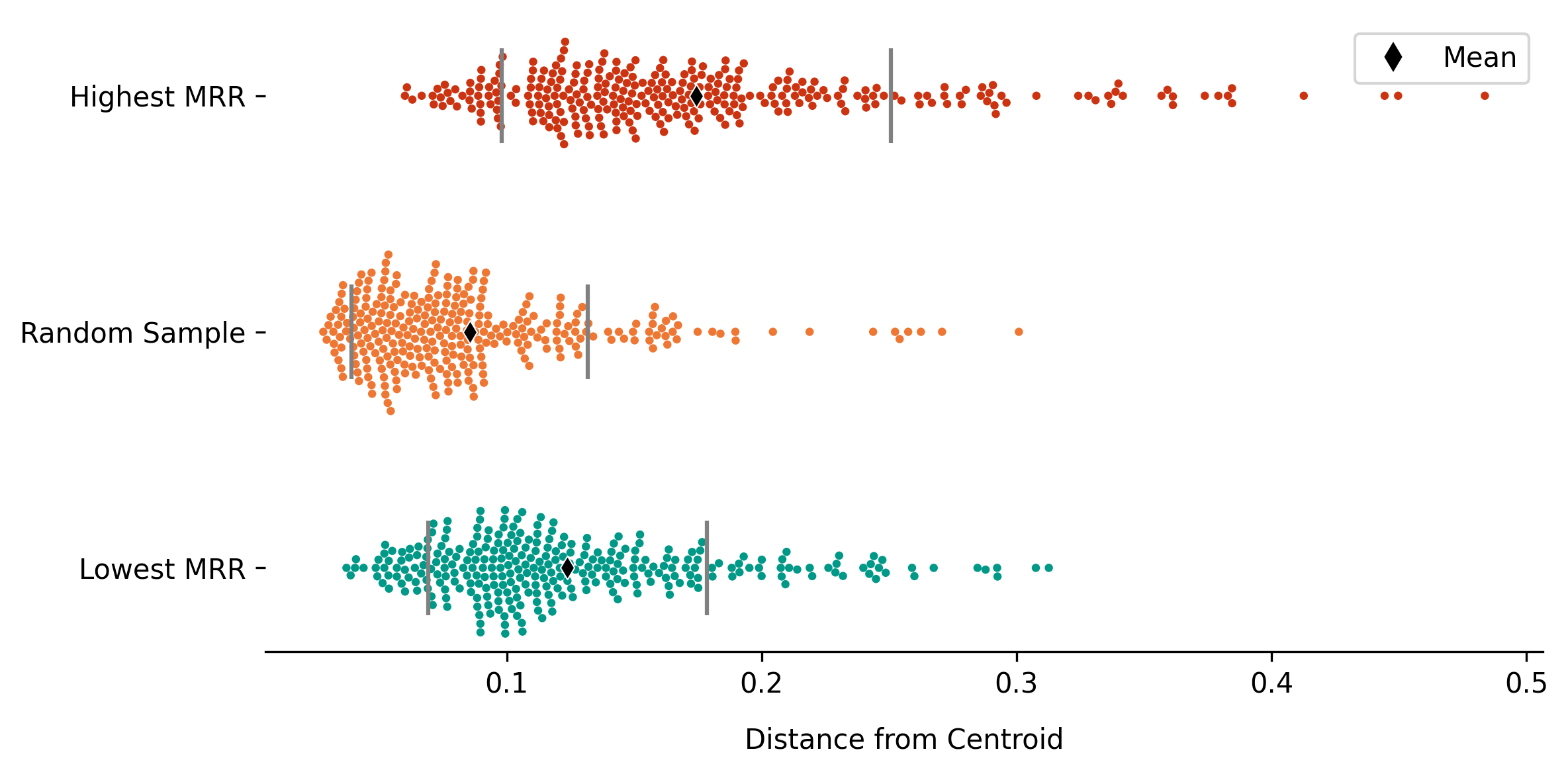}
            \caption{\mpnet}
        \end{subfigure}
        \begin{subfigure}
            [b]{\linewidth}   
            \centering 
            \includegraphics[width=\linewidth]{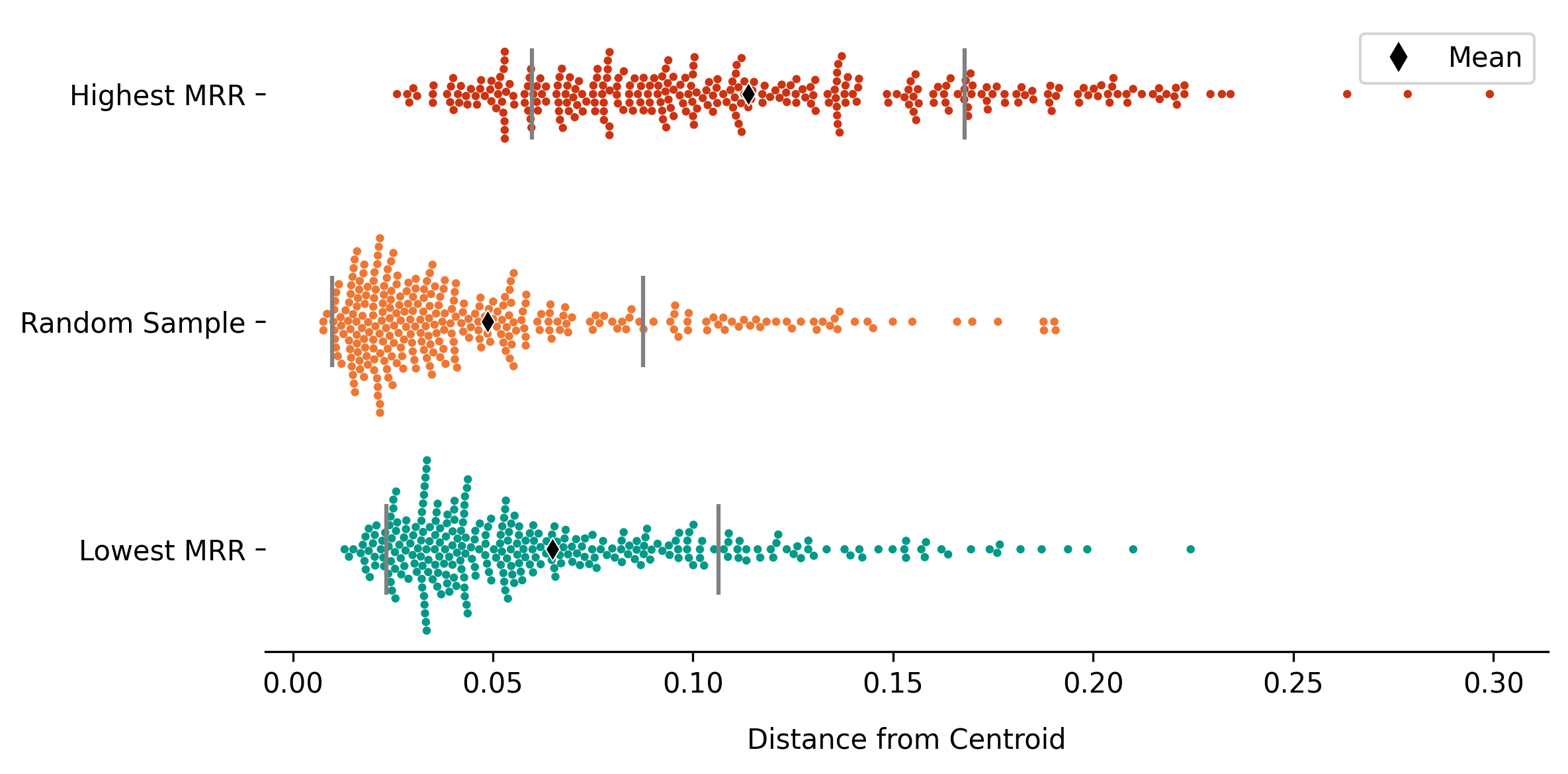}
            \caption{\styledistance}
        \end{subfigure}
        \caption{Reddit - Distribution of Needle Authors' Distances from the Centroid. A comparison between random samples of authors, a group of easily found authors and a group of hard to find authors.}
        \label{fig:needle-dist}
\end{figure}

\begin{figure}[ht!]
    \centering
        \begin{subfigure}[b]{\linewidth}
            \centering
            \includegraphics[width=\linewidth]{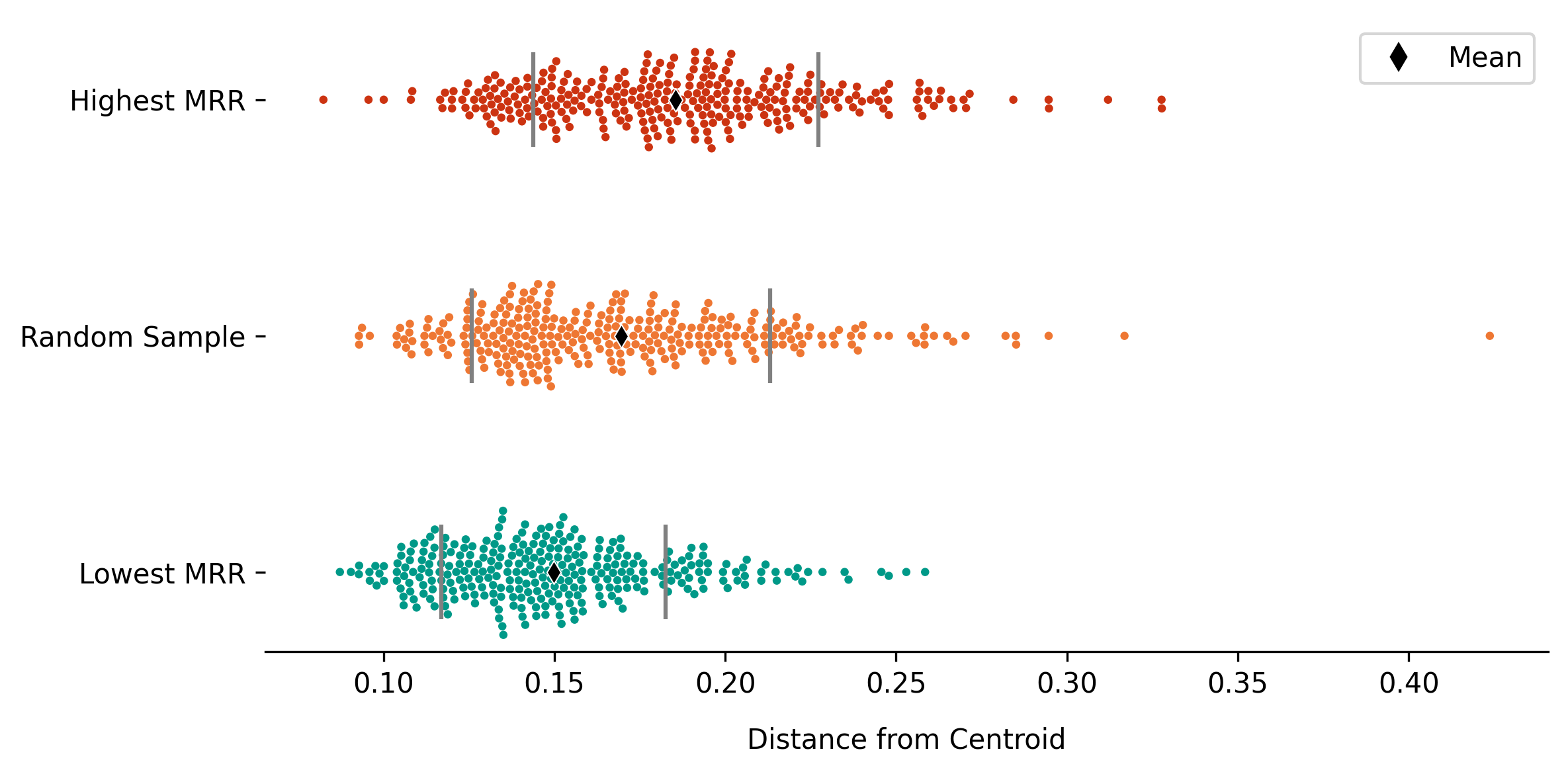}
            \caption{\luar}
        \end{subfigure}
        \begin{subfigure}[b]{\linewidth}
            \centering 
            \includegraphics[width=\linewidth]{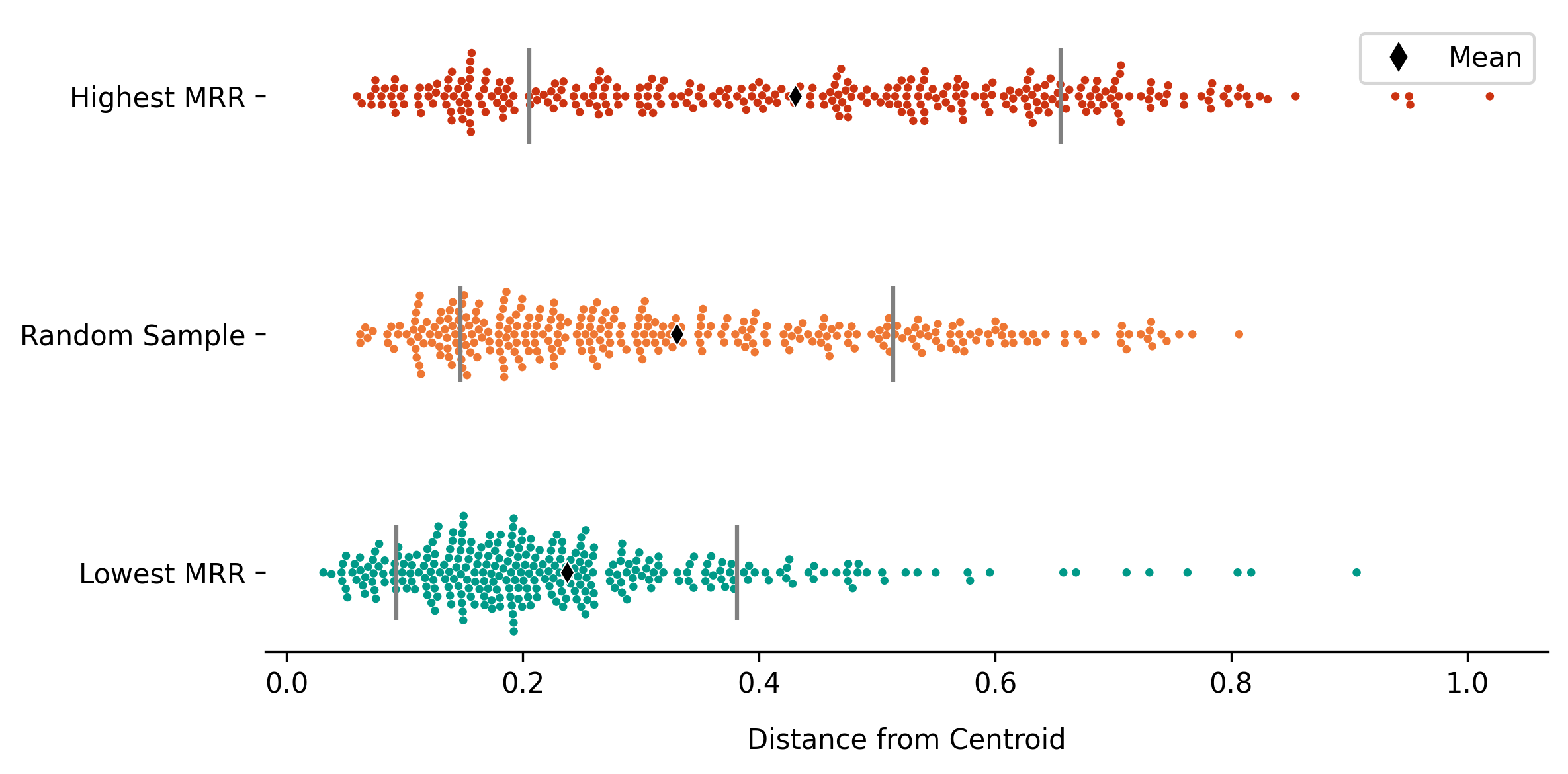}
            \caption{\wegmann}
        \end{subfigure}
        \begin{subfigure}[b]{\linewidth}   
            \centering 
            \includegraphics[width=\linewidth]{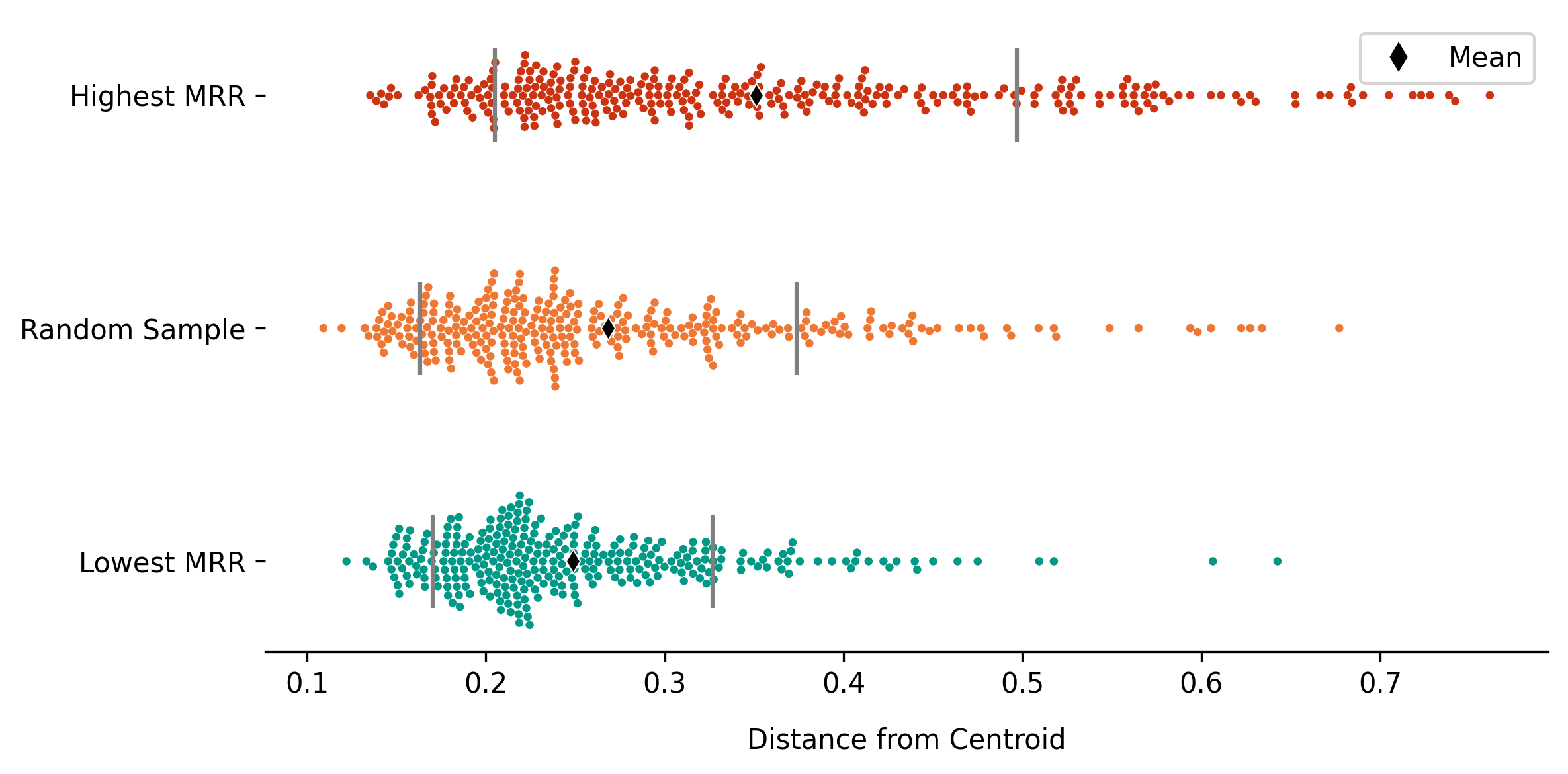}
            \caption{\sbert}
        \end{subfigure}
        \begin{subfigure}[b]{\linewidth}   
            \centering 
            \includegraphics[width=\linewidth]{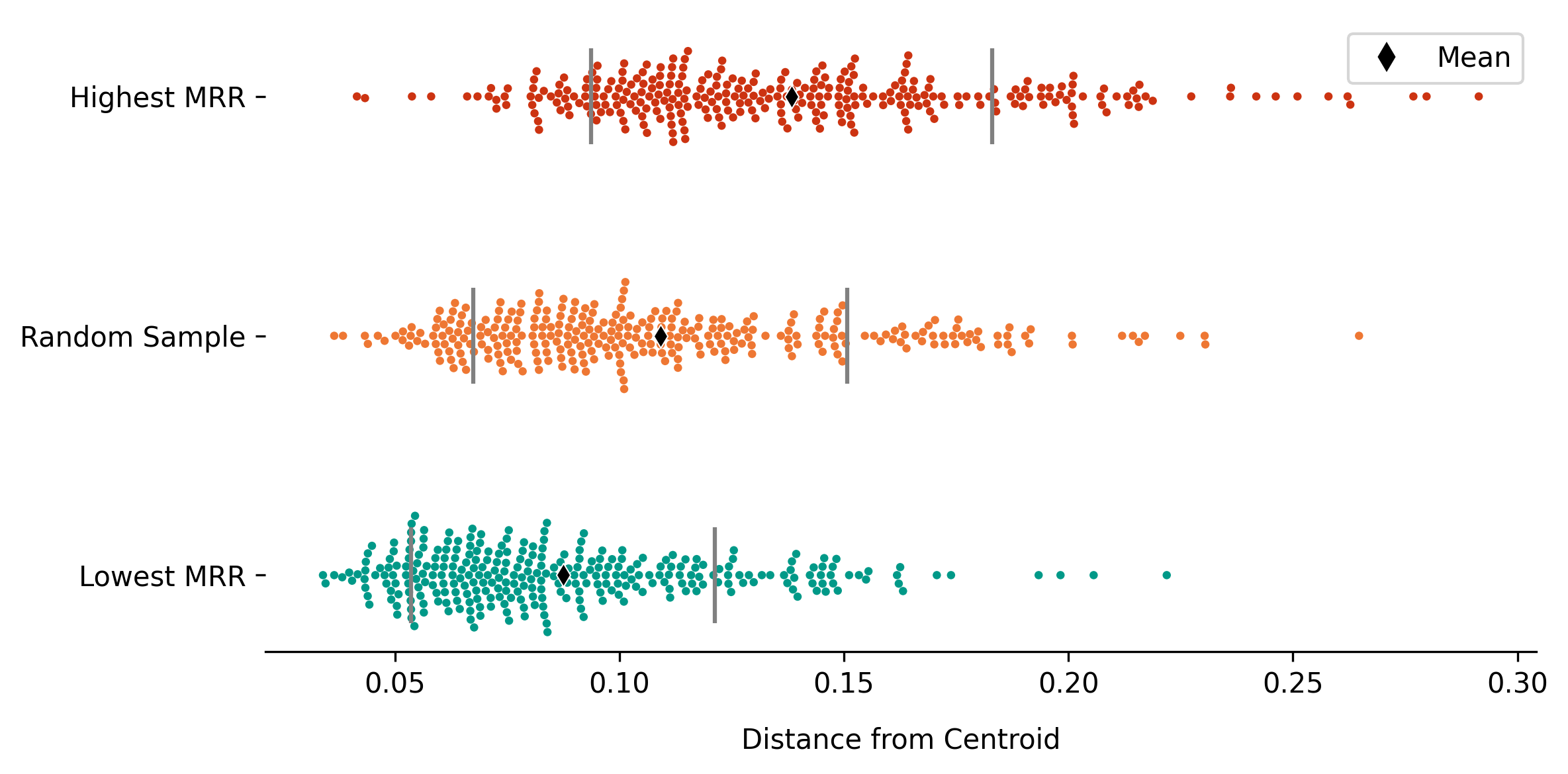}
            \caption{\mpnet}
        \end{subfigure}
                \begin{subfigure}[b]{\linewidth}   
            \centering 
            \includegraphics[width=\linewidth]{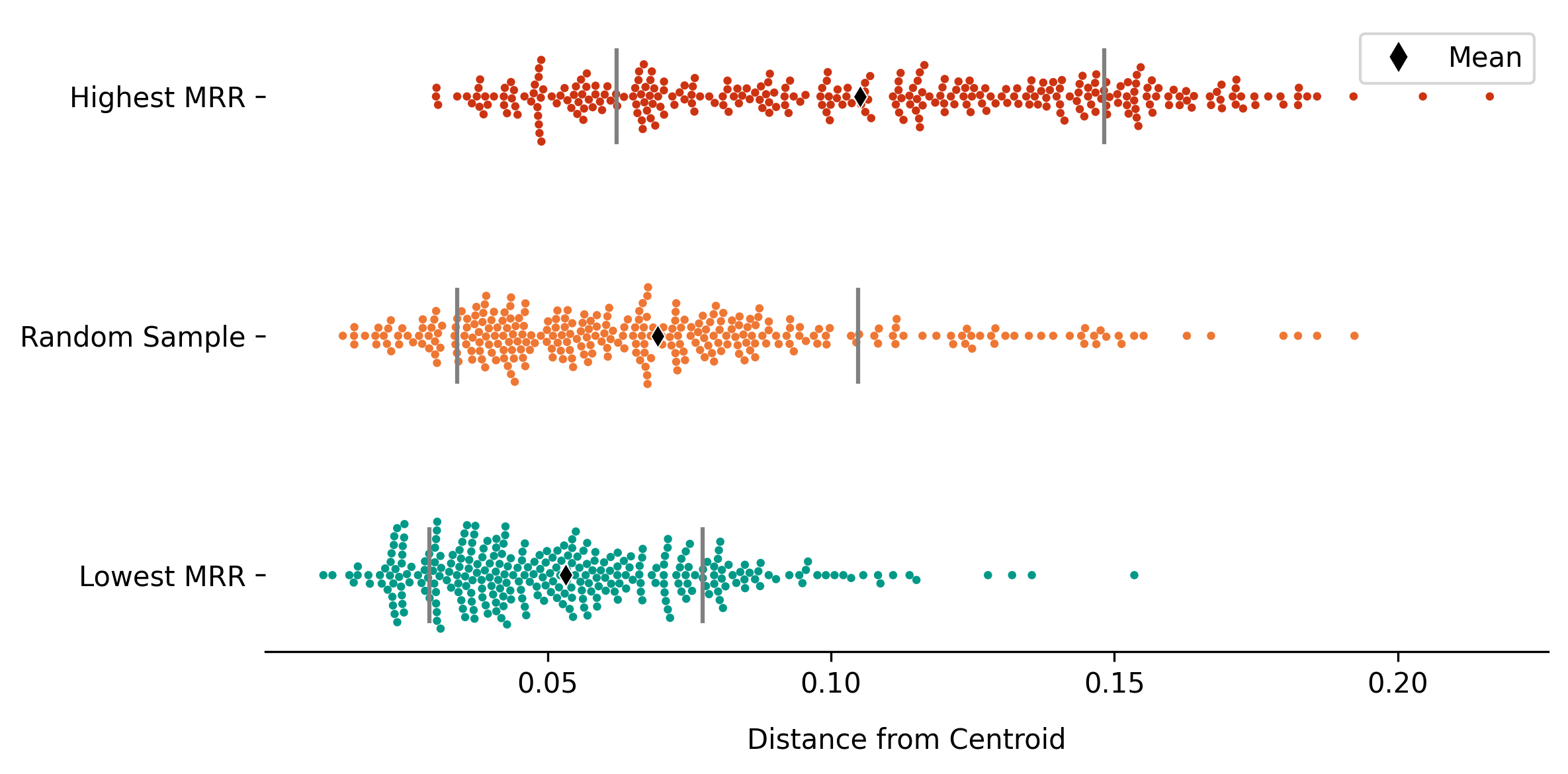}
            \caption{\mpnet}
        \end{subfigure}

        \caption{Blogs - Distribution of Needle Authors' Distances from the Centroid. A comparison between random samples of authors, a group of easily found authors and a group of hard to find authors. }
        \label{fig:blog-needle-dist}
\end{figure}

Test statistics corresponding to section \ref{section-missed-attribution} are presented in \ref{tab:test-stats}. Additionally,
we sample 300 Reddit authors with highest MRRs, 300 authors with lowest MRRs, and 300 random authors to visualize the relationship between needle authors' MRR and their distance from the centroid. Figure \ref{fig:needle-dist} shows the distribution of the sampled authors' distances from the centroid. Similar plots for Blogs experiments are in figure \ref{fig:blog-needle-dist}.

\renewcommand{\arraystretch}{1.1}

\begin{table*}[ht]
\hspace{-35pt}
\smaller
\begin{tabular}{cc|cc|cc|cc|cc|cc|}
\cline{3-12}
 &  & \multicolumn{2}{c|}{\luar} & \multicolumn{2}{c|}{\wegmann} & \multicolumn{2}{c|}{\sbert} & \multicolumn{2}{c|}{\mpnet} & \multicolumn{2}{c|}{\styledistance} \\ \cline{3-12} 
 &  & \multicolumn{1}{c|}{U} & p-value & \multicolumn{1}{c|}{U} & p-value & \multicolumn{1}{c|}{U} & p-value & \multicolumn{1}{c|}{U} & p-value & \multicolumn{1}{c|}{U} & p-value \\ \hline
\multicolumn{1}{|c|}{\multirow{3}{*}{Reddit}} & Hyp. (i) & \multicolumn{1}{c|}{$79199.0$} & $1.12e^{-58}$ & \multicolumn{1}{c|}{$74052.0$} & $6.36e^{-43}$ & \multicolumn{1}{c|}{$59269.0$} & $9.05e^{-12}$ & \multicolumn{1}{c|}{$64956.0$} & $2.75e^{-21}$ & \multicolumn{1}{c|}{$70389.0$} & $2.94e^{-33}$ \\ \cline{2-12} 
\multicolumn{1}{|c|}{} & Hyp. (ii) & \multicolumn{1}{c|}{$71937.5$} & $3.46e^{-37}$ & \multicolumn{1}{c|}{$75528.0$} & $3.51e^{-47}$ & \multicolumn{1}{c|}{$78719.5$} & $4.22e^{-57}$ & \multicolumn{1}{c|}{$79814.5$} & $9.91e^{-61}$ & \multicolumn{1}{c|}{$79633.0$} & $4.03e^{-60}$ \\ \cline{2-12} 
\multicolumn{1}{|c|}{} & Hyp. (iii) & \multicolumn{1}{c|}{$30635.0$} & $6.63e^{-12}$ & \multicolumn{1}{c|}{$50676.0$} & $1.0$ & \multicolumn{1}{c|}{$70489.0$} & $1.0$ & \multicolumn{1}{c|}{$66651.5$} & $1.0$ & \multicolumn{1}{c|}{$62192.5$} & $1.0$ \\ \hline
\multicolumn{1}{|c|}{\multirow{3}{*}{Blogs}} & Hyp. (i) & \multicolumn{1}{c|}{$67607.0$} & $8.91e^{-27}$ & \multicolumn{1}{c|}{$67255.0$} & $5.22e^{-6}$ & \multicolumn{1}{c|}{$65060.0$} & $1.72e^{-21}$ & \multicolumn{1}{c|}{$74715.0$} & $8.26e^{-45}$ & \multicolumn{1}{c|}{$76030.0$} & $1.12e^{-48}$ \\ \cline{2-12} 
\multicolumn{1}{|c|}{} & Hyp. (ii) & \multicolumn{1}{c|}{56918.5} & $9.91e^{-9}$ & \multicolumn{1}{c|}{55346.5} & $5.49e^{-7}$ & \multicolumn{1}{c|}{60090.5} & $5.9e^{-13}$ & \multicolumn{1}{c|}{61172.5} & $1.30e^{-14}$ & \multicolumn{1}{c|}{63759.5} & $4.97e^{-19}$ \\ \cline{2-12} 
\multicolumn{1}{|c|}{} & Hyp. (iii) & \multicolumn{1}{c|}{$34128.0$} & $1.52e^{-7}$ & \multicolumn{1}{c|}{$31362.0$} & $6.66e^{-11}$ & \multicolumn{1}{c|}{$41619.0$} & $0.055$ & \multicolumn{1}{c|}{$29477.0$} & $1.33e^{-13}$ & \multicolumn{1}{c|}{$30604.5$} & $6e^{-12}$ \\ \hline
\end{tabular}
\caption{Mann-Whitney U-statistics and p-values of testing three hypotheses about our authorship attribution experiments: (i) Authors with higher MRR are further away from the centroid compared to authors with lower MRR, (ii) Authors with higher MRR are further away from the centroid compared to a random subset of authors. (iii) Authors with lower MRR are closer to the center than a random subset of authors.}
\label{tab:test-stats}

\end{table*}

\section{Risk of Misattribution for Different Authors}
\label{apx:risk_across_authors}

Not all authors who are subject to over-misattribution carry the same risk. Per dataset and model, we obtain the ratio of the number of times an author is ranked in top 10 to $E_{10}$: $u^{10}_j = \frac{c^{10}_j}{E_{10}}$. For all authors contributing to the \ufar{10} score (eq. \ref{ufar}) this ratio is more than 1. Table \ref{tab:num-exceed} shows for how many authors this ratio is higher than 2, 4, and 5. 
Table \ref{tab:risk-across-authors} shows the maximum, mean, and standard deviation of $u^{10}$ across authors. 
\renewcommand{\arraystretch}{1}
\end{document}